\pdfoutput=1

\documentclass[11pt,table,dvipsnames]{article}

\usepackage{acl}
\usepackage{times}
\usepackage{latexsym}
\usepackage[T1]{fontenc}

\usepackage[utf8]{inputenc}

\usepackage{mdframed}

\usepackage{microtype}
\usepackage{latexsym}
\usepackage{dsfont}
\usepackage{booktabs} 
\usepackage{subfigure}
\usepackage{graphicx}
\usepackage{bigstrut}
\usepackage{multirow}
\usepackage{floatrow}
\usepackage{algpseudocode}
\usepackage[normalem]{ulem}
\usepackage{amssymb}
\usepackage{amsfonts}
\usepackage{multicol}
\usepackage{tipa}
\usepackage[ruled,linesnumbered]{algorithm2e}

\usepackage{paralist}
\usepackage[switch]{lineno}
\usepackage{multirow, makecell, caption}
\usepackage{arydshln}
\usepackage{verbatim}

\usepackage{csquotes}
\usepackage{xspace}
\usepackage{paralist}
\usepackage{mdwlist}
\usepackage{subfigure}
\usepackage{array}
\usepackage{enumitem}
\usepackage{bm}
\usepackage{colortbl}
\usepackage{dsfont}
\usepackage{url}
\usepackage{graphicx}
\usepackage{tabularx}
\usepackage{amsmath}
\usepackage{bbm}
\usepackage{tikz}
\usepackage{pgfplots}

\newcommand{\method}{\textsc{Segment\textsuperscript{+}}\xspace}

\newcommand{\mrcolorOne}{\cellcolor[rgb]{0.902,0.957,0.945}}
\newcommand{\mrcolorTwo}{\cellcolor[rgb]
{0.976, 0.984, 0.906}}

\newcommand{\mrcolorThree}{\cellcolor[rgb]{0.922, 0.992, 1}}
\newcommand{\mrcolorFour}{\cellcolor[rgb]{0.922, 0.933, 1}}
\newcommand{\mrcolorFive}{\cellcolor[rgb]{0.910, 0.970, 1}}
\newcommand{\mrcolorSix}{\cellcolor[rgb]
{0.940, 0.940, 0.940}}

\makeatletter
\def\adl@drawiv#1#2#3{%
        \hskip.5\tabcolsep
        \xleaders#3{#2.5\@tempdimb #1{1}#2.5\@tempdimb}%
                #2\z@ plus1fil minus1fil\relax
        \hskip.5\tabcolsep}
\newcommand{\cdashlinelr}[1]{%
  \noalign{\vskip\aboverulesep
           \global\let\@dashdrawstore\adl@draw
           \global\let\adl@draw\adl@drawiv}
  \cdashline{#1}
  \noalign{\global\let\adl@draw\@dashdrawstore
           \vskip\belowrulesep}}
\makeatother

\title{\method: Long Text Processing with Short-Context Language Models}

\author{\parbox{\textwidth}{\centering
Wei Shi\textsuperscript{\rm $\spadesuit$}, 
Shuang Li\textsuperscript{\rm $\spadesuit$}, 
Kerun Yu\textsuperscript{\rm $\heartsuit$}, \\
Jinglei Chen\textsuperscript{\rm $\clubsuit$},  
Zujie Liang\textsuperscript{\rm $\clubsuit$}, 
Xinhui Wu\textsuperscript{\rm $\clubsuit$},
Yuxi Qian\textsuperscript{\rm $\clubsuit$}, 
Feng Wei\textsuperscript{\rm $\clubsuit$}, 
Bo Zheng\textsuperscript{\rm $\clubsuit$},\\ 
Jiaqing Liang\textsuperscript{\rm $\diamondsuit$}, 
Jiangjie Chen\textsuperscript{\rm $\spadesuit$\thanks{~~Corresponding authors.}}, 
Yanghua Xiao\textsuperscript{\rm $\spadesuit$}\footnotemark[1]
}\\
\textsuperscript{\rm $\spadesuit$}Shanghai Key Laboratory of Data Science, School of Computer Science, Fudan University\\
\textsuperscript{\rm $\heartsuit$}Columbia University \quad
\textsuperscript{\rm $\clubsuit$}MYbank, Ant Group\\
\textsuperscript{\rm $\diamondsuit$}School of Data Science, Fudan University\\
\texttt{wshi22@m.fudan.edu.cn}\quad
\texttt{\{jjchen19, shawyh\}@fudan.edu.cn}
}
\DeclareUnicodeCharacter{266B}{}
\begin{document}

\maketitle

\begin{abstract}
There is a growing interest in expanding the input capacity of language models (LMs) across various domains. However, simply increasing the context window does not guarantee robust performance across diverse long-input processing tasks, such as understanding extensive documents and extracting detailed information from lengthy and noisy data. In response, we introduce \method, a general framework that enables LMs to handle extended inputs within limited context windows efficiently. \method utilizes structured notes and a filtering module to manage information flow, resulting in a system that is both controllable and interpretable. Our extensive experiments across various model sizes, focusing on long-document question-answering and Needle-in-a-Haystack tasks, demonstrate the effectiveness of \method in improving performance.\footnote{Our code
is available at \url{https://github.com/WeiShi-9/segmentplus}.}
\end{abstract}

\section{Introduction}
\label{sec:intro}
Language models (LMs) have shown exceptional performance in a wide range of NLP tasks~\cite{pu2023summarization, wei2022emergent, NEURIPS2022_9d560961}.
Due to the relatively short context window of most LMs, they face unique challenges in contexts such as long-document question answering, long-term memory maintenance, and processing lengthy, noisy contexts~\cite{shaham-etal-2022-scrolls, bai2023longbench,packer2023memgpt,liu2023lost, LLMTest_NeedleInAHaystack2023}. Efficiently processing long texts across various tasks remains a core challenge in this community.

\begin{figure}[t!]
    \centering
    \includegraphics[width=\linewidth]{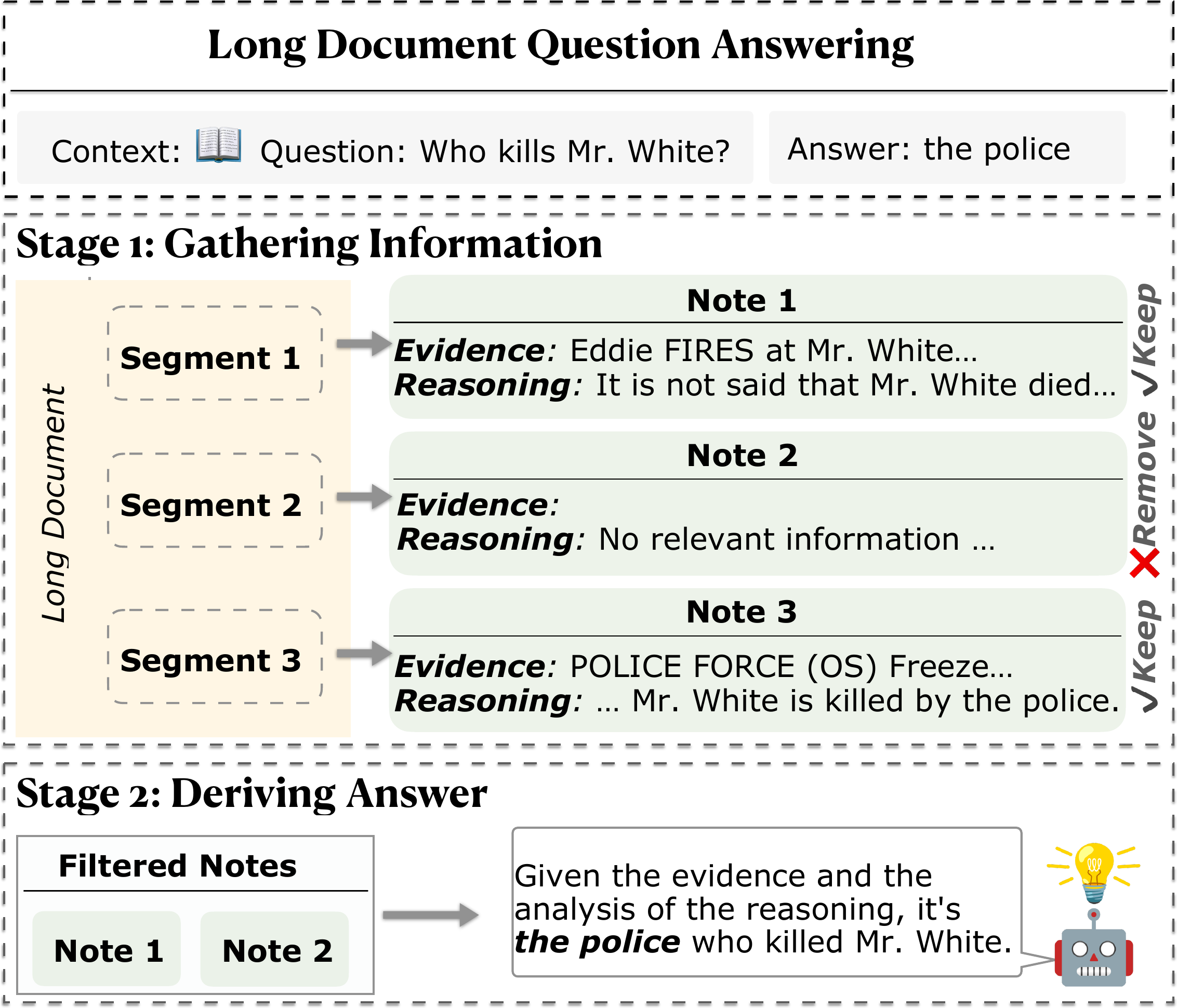}
    \caption{This picture illustrates the use of short-context models to tackle long document question answering tasks in \method. The process begins by gathering relevant context from the document for a specific question. Only notes labeled 'keep' are used as the context to derive the final answer, avoiding noise.}
    \label{fig:front}
\end{figure}

To reduce input length for handling long text, traditional retrieval is a simple and fast method but struggles with tasks requiring multiple pieces of information, often missing details and introducing noise~\cite{wang2023learning}. Enhancements like query rewriters~\cite{ma-etal-2023-query,chan2024rqrag} and feedback loops~\cite{asai2023selfrag} have reduced its usability and tied it more closely to specific downstream tasks.
For handling long inputs at once, a series of works focus on broadening the context window of language models~\cite{touvron2023llama, Douzon_2023}.
This method also has intrinsic limitations: the context window cannot be infinitely expanded, and these models cannot maintain robustness across various tasks and context lengths.~\cite{he2023lost,LLMTest_NeedleInAHaystack2023,hsieh2024ruler}.

To access the entire text and perform multiple intermediate reasoning steps to enhance performance on complex tasks, memory management involves repeatedly using short-context LMs combined with document pre/post-processing for managing long texts~\cite{chen2023walking, packer2023memgpt}.
For such systems, the key challenge lies in designing a mechanism to manage information flow across different invocation times.
However, past research often relies heavily on the model's inherent capabilities for planning and spontaneous decision-making, resulting in an uncontrolled reasoning process with noisy, free-form text expressions.

To address the above issues and challenges, we propose the \textbf{\method}, a robust and controllable framework that helps language models process long texts with a limited context window.
The key insight of the \method is capturing the characteristics of queries and designing two specific components to gather and merge different types of information from long inputs. 
As shown in Figure ~\ref{fig:front}, the \method agent has two stages. In stage 1, the model collects structured notes in parallel from all segments, with each note containing an Evidence and a Reasoning part. After filtering out unhelpful notes labeled as `Remove', the remaining notes proceed to stage 2. In stage 2, the notes are divided into batches, maintaining the same order as the input, with a maximum token limit. Each batch of notes is then merged into one updated structured note. This process is iterated until the remaining notes can fit into the context window as the final context for answering the question.

The challenge of processing long texts can be effectively tackled by using two types of components for information flow control. We notice that some questions require specific detailed information, while others need further reasoning across different parts of the content. Therefore, we create an Evidence component for gathering original sentences from the input, focusing on precision, and a Reasoning component to help the model compress context into high-level semantic information, focusing on recall. This division, using both Evidence and Reasoning, makes the process both controllable and interpretable.

Significantly, retrieval and long-context language models can both benefit our method by either moderately increasing the processing context window or narrowing the input range that needs to be processed, which does not require high accuracy.

In short, our contributions include:
\begin{inparaenum}[\it 1)]
    \item We introduce a versatile framework for long context processing, applicable across language models of varying sizes and multiple text domains.
    \item Our method, leveraging a robust reasoning schema, outperforms other agent-based baselines and advanced long context models in long text processing tasks.
    \item We conduct a thorough analysis of \method, highlighting the importance of structured information control.
\end{inparaenum}

\section{Related Work}
\label{sec:related}
\paragraph{Retrieval-augmented Generation}

With a dense or sparse retriever, we can swiftly find relevant information in long texts by comparing query similarity.
However, direct use of user queries to retrieve relevant information may not always yield useful results due to ambiguity or incomplete queries~\cite{ma2023query,liu2023retallm}, thereby introducing noise~\cite{wang2023learning}.
For re-managing the retrieved data, ~\citet{zhu2023large, zhuang-etal-2023-open} explore the information organization capabilities of language models, utilizing LLMs as rerankers for more precise sorting.
While single-turn retrieval may bring in limited useful information, insufficient for some queries, some studies ~\cite{jiang2023active, shao2023enhancing} focus on multiple searches based on language model outputs, which may yield superior results. 
Despite these efforts, the information retrieved often remains fragmented, incomplete, and only partially represents the original materials from the long input. This fragmentation presents challenges for tasks requiring the synthesis and reasoning across multiple segments of a long text~\cite{behnamghader-etal-2023-retriever}. Our method addresses this issue by using structured information gathering that includes not only the essential original evidence but also segment-aware analysis for further explanation, thus facilitating easier reasoning over the entire long input.

\paragraph{Long Context LMs}
Language models perform well in a variety of applications but struggle with large texts due to limited context windows~\cite{shaham-etal-2022-scrolls,bai2023longbench,packer2023memgpt}. 
Through techniques such as position interpolation and continuous pretraining, researchers have attempted to expand the context windows, thereby improving performance for both long and short document tasks~\cite{chen2023extending, xiong2023effective}. 
However, these approaches are limited by data quality and feasible window size constraints~\cite{xiong2023effective}. 
Besides, the models' inability to handle queries when key information is scattered across a large text is also a notable challenge~\cite{liu2023lost}.~\citet{he2023lost} discovers that this issue arises from attention failure and can be alleviated by training models with a specially designed task. 
Thus, the performance and robustness of models processing very long texts in a longer context remain uncertain, and limited resources restrict our ability to indefinitely expand the model's context window.
Our method addresses this challenge by dynamically optimizing the use of the available context window, thus allowing for the effective extraction of crucial information scattered throughout the text. This approach shifts the focus from merely widening the window to better utilizing the model’s current context capabilities.

\paragraph{Memory Management}
Rather than processing long text in one go, we can use language models as agents to handle the long input task step by step.
In such systems, memory is pivotal not only for storing out-of-window information but also as a foundation for lifelong learning through historical data analysis~\cite{sumers2023cognitive,majumder2023clin}. 
Using LLM agents for long text tasks has its advantages.
Firstly, efficiently organizing and utilizing memory can improve performance in both documents QA and dialogue tasks~\cite{packer2023memgpt}.
Secondly, capturing the document structure and employing agent navigators across document segments is advantageous for document QA tasks~\cite{chen2023walking}.
Lastly, leveraging the task decomposition and plan-and-solve abilities of agents also benefits these tasks~\cite{sun2023pearl}.
Our approach stands out by showcasing how it's effective across variously sized models, unlike others that require high language model capabilities.


\section{Method}
\label{sec:method}
\begin{figure}[t!]
    \centering
    \includegraphics[width=\linewidth]{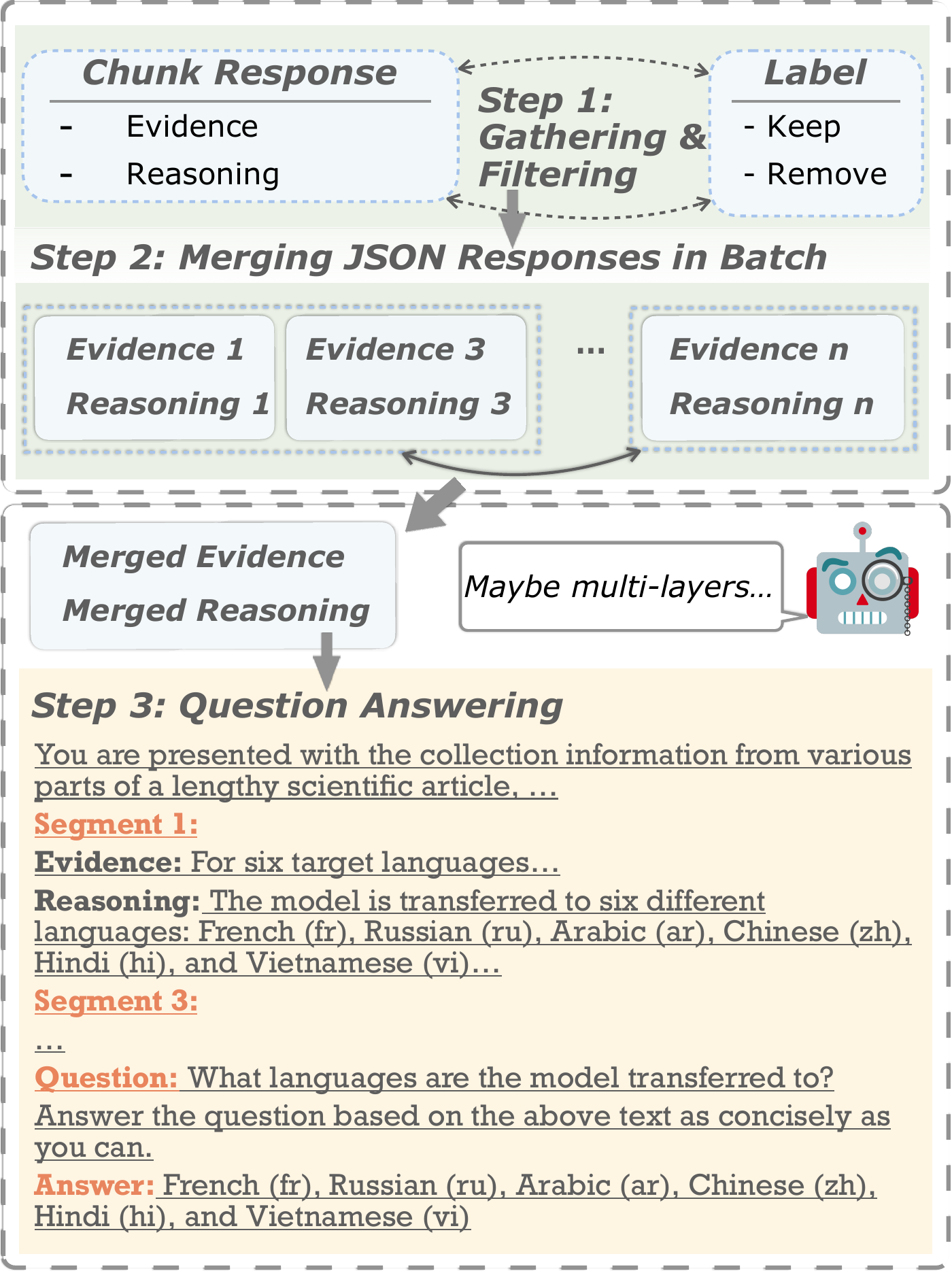}
    \caption{The proposed framework for \method consists of three main components. First, a gathering module collects structural information for a given query, distinguishing direct, accurate context (evidence) from the model’s potentially misleading analysis (reasoning). Next, a filter module filters out noisy segments for dense information management. Finally, we merge this information in batches, taking into account the limited context window of the merging language model, to produce a suitable length context optimized for final answering.}
    \label{fig:method}
\end{figure}

The core challenge of processing long inputs with short-context models is how to control the information flow within different segments.
In other words, how to retain the most useful information while using the fewest tokens. 
We first construct this process into a two-stage pipeline and then introduce each step in detail.

\subsection{Problem Formulation}

First, we can approach the long input processing task as a two-stage process: \textbf{\textit{1) Gathering information}} from different parts of the input, and \textbf{\textit{2) Reasoning over this information}}, performing further inference, and eliminating duplications for the final output.
From this perspective, traditional retrieval techniques directly address the first phase by selecting passages with the highest similarity to the question as context, and then explicitly carry out the second step for reasoning. Similarly, long-context-window models also perform these actions but do so implicitly to provide an answer. In contrast, our proposed \method clearly and efficiently performs these two actions, providing a more transparent and manageable information flow.

\subsection{\method}

The first question we focus on is \textbf{\textit{how to efficiently and losslessly collect all the useful information from segments}}. Before addressing this problem, we observe that common questions about long inputs can be divided into two types~\citep{li2023loogle,pang-etal-2022-quality}: short-dependency questions and long-dependency questions. Short-dependency questions may only require several sentences or words for the final answer and need direct and exact information. Long-dependency questions, on the other hand, require the agent to aggregate information from different parts of the input and perform reasoning for the final answer, necessitating comprehensive and concisely compressed information. 

Motivated by these data characteristics, we designed a specific structure named \textit{\textbf{Note}} for gathering information that contains two components: \textbf{\textit{Evidence}} and \textbf{\textit{Reasoning}}, which together form a set of notes. The Evidence part requires the model to collect original sentences from one segment that directly answer the asked question, corresponding to short-dependency questions and improving the precision of the information. The Reasoning part requires the model to gather possible information and compress it into concise texts related to the question, such as mentioned entities and events, corresponding to long-dependency questions and improving the recall of the information. Just like human reading habits, we may underline key sentences in the inputs and make annotations to aid further reasoning tasks.

In the first step, given a question \textit{\(Q\)} and a list of potentially useful segments \textit{\(C = \{c_1, c_2, \ldots, c_n\}\)}, for each segment \textit{\(c_i\)}, we create a \textit{Note}, which is processed in parallel.
We define this process as shown in Figure ~\ref{fig:method}: \textbf{\textit{1)}} For each segment \(i\), we collect structured information relative to the question, represented as a JSON object: \[ note_i = \{``Evidence": ``", ``Reasoning": ``"\} \] 

We then filter the notes to keep them information-dense and remove noise. This is crucial for small models that lack robustness. When dealing with short-dependency questions, 10 segments produce 10 notes, but only one contains useful information. The others might state "No information" or the model might generate a possible but irrelevant answer. In such cases, the model might fail to identify the real useful information and could be influenced by crowd psychology or noise~\citep{xie2024adaptivechameleonstubbornsloth}. In our well-structured notes, this problem can be mitigated by letting the model filter out notes with empty evidence or those that clearly state "no information" in the reasoning part. We then keep all the other notes, even if there is redundant information.
As shown in Figure ~\ref{fig:method}: \textbf{\textit{1)}}, each \( note_i \) is labeled by a filter, denoted as \( F \), with either a `Keep' or `Remove' label. We prompt language models with the question and JSON responses, evaluating their relevance based on whether they provide at least one piece of useful information. Segments that lack relevant information are tagged for removal to prevent them from leading the model toward an undesired outcome.
This component can be trained using task-specific data or optimized through prompt engineering.

In the second step, the core question is \textbf{\textit{how to best maintain the gathered information with the least tokens and without loss}}. This can be addressed with a simple solution in our structured information flow control. We first keep the collected parallel notes in the same order as the segments to maintain some semantic information. Before merging these notes, we split them into batches based on a given max token limit due to the model's context window constraints. In each batch, we concatenate the evidence parts directly, as they contain exact useful information. For the reasoning part, the model performs further inference and refines redundant information. This way, each batch yields a new merged note that maintains the structure.
This process can be repeated over multiple iterations to generate a final merged note that fits within the model's context window limit.

In the third step, to get the final answer, we use the final merged Notes as the context for model prediction. In other words, the model does not interact with the original segments.

The reasoning process in our designed framework is bottom-up.
Our information collection focuses on retrieving data with a reasoning process that directly related to the final question and parts of the question or subquestions, rather than relying solely on semantic matching for information retrieval.
This approach is useful for addressing both one-hop and multi-hop questions and tasks requiring knowledge synthesis from different segments. 
For example, in a two-hop question like \textit{``The rapper who owns Aspiro was inspired by what when writing Song Cry?"}, the model gathers information on both the rapper's identity and songwriting experience from each segment and performs reasoning during the merging process or final answer generation. This process can reduce costs and improve efficiency compared to multi-turn interaction searching and communications.

In conclusion, these design concepts assure a strong and efficient framework for providing accurate and comprehensive answers in the long text processing domain, including organized information collecting, strategic response labeling, and a focus on partial information gathering to solve complicated, multifaceted queries.

\section{Experiments}
\label{sec:experiments}
In this section, we apply the \method to two distinct long text reasoning tasks: long document question answering and needle-in-a-haystack tasks.
For the long document QA task, we analyze \method's ability to compress reading contexts and merge information efficiently in parallel (§\ref{sec:longdoc}).
For the needle-in-a-haystack task, we evaluate \method's resistance to noise while accurately gathering essential sentences and do reasoning for final answer.(§\ref{sec:babilong}).

\subsection{Long Document Question Answering}
\label{sec:longdoc}
\begin{table}[ht]
\centering
\resizebox{\textwidth}{!}{%
\begin{tabular}{@{}lccccc@{}}
\toprule
& Qasper & MSQ & HQA & NQA & QLTY \\ \midrule
\text{Max} & 19372 & 16337 & 16325 & 476004 & 8609 \\
\text{Min} & 1785 & 6484 & 1748 & 8961 & 2401 \\
\text{Avg} & 4880.54 & 15576.98 & 12793.29 & 75678.19 & 5613.78 \\ \bottomrule
\end{tabular}%
}
\caption{Summary of the maximum (Max), minimum (Min), and average (Avg) token counts for selected long document question answering datasets, tokenized using ChatGPT's tokenizer.}
\label{tab:selected_dataset_tokens}
\end{table}

{
\begin{table*}[t!]
\centering
\begin{tabular}{@{}cccccccccccc@{}}

\toprule
\multirow{2}{*}{\textbf{Model}} & \multirow{2}{*}{\textbf{Method}} & \multicolumn{2}{c}{\textbf{Qasper}} & \multicolumn{2}{c}{\textbf{MSQ}} & \multicolumn{2}{c}{\textbf{HQA}} & \multicolumn{2}{c}{\textbf{NQA}} & \textbf{QLTY} & \textbf{Avg.}\\
\cmidrule(lr){3-4} \cmidrule(lr){5-6} \cmidrule(lr){7-8} \cmidrule(lr){9-10} \cmidrule(lr){11-11} \cmidrule(lr){12-12}
&                         & \textbf{GPT4}         & \textbf{Auto}         & \textbf{GPT4}          & \textbf{Auto}         & \textbf{GPT4}          & \textbf{Auto}          & \textbf{GPT4}        & \textbf{Auto}  & \textbf{GPT4} & \textbf{GPT4}   \\
\midrule

\multirow{6}{*}{ChatGPT}    & MemGPT (16k)   & \mrcolorOne-     & \mrcolorOne-     & \mrcolorTwo-     & \mrcolorTwo-     & \mrcolorThree-     & \mrcolorThree-     & \mrcolorFour-     & \mrcolorFour-     & \mrcolorFive-         & \mrcolorSix   -   \\
& Pearl (16k)    & \mrcolorOne51.46 & \mrcolorOne31.19 & \mrcolorTwo26.47 & \mrcolorTwo9.06  & \mrcolorThree38.55 & \mrcolorThree12.85 & \mrcolorFour20.57 & \mrcolorFour5.83  & \mrcolorFive64.63     & \mrcolorSix40.34 \\
& Vanilla (4k)  & \mrcolorOne47.75 & \mrcolorOne31.91 & \mrcolorTwo23.67 & \mrcolorTwo24.30  & \mrcolorThree54.55 & \mrcolorThree45.24 & \mrcolorFour29.95 & \mrcolorFour21.57 &   \mrcolorFive70.00        & \mrcolorSix45.18 \\
& Retrieve (4k) & \mrcolorOne46.80  &  \mrcolorOne33.36 & \mrcolorTwo26.40  & \mrcolorTwo26.27 & \mrcolorThree63.12 &   \mrcolorThree53.56 & \mrcolorFour32.50  & \mrcolorFour21.28 & \mrcolorFive73.50      & \mrcolorSix48.46 \\
 & Long (16k)  & \mrcolorOne54.10 &\mrcolorOne \textbf{36.78}&\mrcolorTwo\textbf{39.93} &	\mrcolorTwo\textbf{37.11} & \mrcolorThree\textbf{68.45}&	\mrcolorThree56.30 &\mrcolorFour38.45 & \mrcolorFour26.78 &\mrcolorFive74.50 &       \mrcolorSix55.09 \\
& Segment\textsuperscript{+} (4k)     &   \mrcolorOne\textbf{56.29} & \mrcolorOne25.66 &   \mrcolorTwo36.38 &   \mrcolorTwo30.81    &   \mrcolorThree67.96 & \mrcolorThree\textbf{56.76} &   \mrcolorFour\textbf{42.00}  & \mrcolorFour \textbf{26.88}&\mrcolorFive\textbf{75.00}        & \mrcolorSix\textbf{55.53} \\
\cdashlinelr{1-12}
\multirow{6}{*}{GPT-4}      & MemGPT (128k)   & \mrcolorOne55.90  & \mrcolorOne22.60  & \mrcolorTwo39.58 & \mrcolorTwo33.42 & \mrcolorThree67.90  & \mrcolorThree50.03 & \mrcolorFour48.21 & \mrcolorFour19.15 & \mrcolorFive74.47     & \mrcolorSix57.21 \\
& Pearl (128k)    &   \mrcolorOne61.03 & \mrcolorOne36.01 & \mrcolorTwo40.13 & \mrcolorTwo12.25 & \mrcolorThree64.77 & \mrcolorThree18.47 & \mrcolorFour38.38 & \mrcolorFour10.00    & \mrcolorFive81.92     & \mrcolorSix57.25 \\
& Vanilla (16k)  & \mrcolorOne51.38 &   \mrcolorOne36.33 & \mrcolorTwo27.63 & \mrcolorTwo26.92 & \mrcolorThree55.88 & \mrcolorThree47.56 & \mrcolorFour37.90  & \mrcolorFour22.64 &   \mrcolorFive77.00        & \mrcolorSix49.96  \\
& Retrieve (4k) & \mrcolorOne51.98 & \mrcolorOne35.47 & \mrcolorTwo34.15 & \mrcolorTwo32.27 &   \mrcolorThree70.96 & \mrcolorThree59.06 & \mrcolorFour50.00  &   \mrcolorFour\textbf{50.00}   &  \mrcolorFive 83.50      & \mrcolorSix58.12 \\
& Long (16k)     &  \mrcolorOne54.72  & \mrcolorOne \textbf{38.54}    &  \mrcolorTwo\textbf{51.15}  &  \mrcolorTwo \textbf{50.53}   &  \mrcolorThree79.07  &  \mrcolorThree \textbf{67.82}   &  \mrcolorFour41.50   &   \mrcolorFour26.31    &  \mrcolorFive\textbf{90.50 }  &  \mrcolorSix        63.39  \\
& Segment\textsuperscript{+} (4k)     & \mrcolorOne\textbf{63.52} & \mrcolorOne 25.37&   \mrcolorTwo48.82 &  \mrcolorTwo44.97  & \mrcolorThree\textbf{80.00} & \mrcolorThree65.79   & \mrcolorFour\textbf{54.45} & \mrcolorFour30.86  &  \mrcolorFive 88.50        & \mrcolorSix \textbf{67.06} \\
\cdashlinelr{1-12}
\multirow{4}{*}{Vicuna-7B}  & Vanilla (4k) & \mrcolorOne35.65 & \mrcolorOne20.33 & \mrcolorTwo12.38 & \mrcolorTwo6.23  & \mrcolorThree38.85 & \mrcolorThree21.69 & \mrcolorFour12.05 & \mrcolorFour9.14  &  \mrcolorFive 37.50        & \mrcolorSix27.29 \\
& Retrieve (4k) &  \mrcolorOne 39.75 &   \mrcolorOne\textbf{24.15} & \mrcolorTwo17.48 & \mrcolorTwo\textbf{8.51}  &   \mrcolorThree\textbf{46.17} & \mrcolorThree\textbf{23.84} & \mrcolorFour\textbf{25.61} &   \mrcolorFour\textbf{16.62} & \mrcolorFive33.00 & \mrcolorSix32.40 \\
& Long (16k)     &  \mrcolorOne30.14  & \mrcolorOne   21.36  &  \mrcolorTwo13.90  & \mrcolorTwo  7.43    &  \mrcolorThree43.98  &  \mrcolorThree 22.69    &  \mrcolorFour20.30   & \mrcolorFour 12.59     &  \mrcolorFive 40.00  & \mrcolorSix29.66         \\
& Segment\textsuperscript{+} (4k)     & \mrcolorOne\textbf{39.80} & \mrcolorOne14.94 &   \mrcolorTwo\textbf{19.00} &   \mrcolorTwo8.19 & \mrcolorThree44.42 &   \mrcolorThree19.55 &   \mrcolorFour23.85 & \mrcolorFour11.52 & \mrcolorFive\textbf{46.00}     & \mrcolorSix\textbf{34.61}\\

\cdashlinelr{1-12}
\multirow{4}{*}{Vicuna-13B} & Vanilla (4k) & \mrcolorOne29.50  & \mrcolorOne18.35 & \mrcolorTwo16.38 & \mrcolorTwo13.04 & \mrcolorThree42.82 & \mrcolorThree30.02 & \mrcolorFour22.90  & \mrcolorFour13.77 &  \mrcolorFive42.00         & \mrcolorSix30.72  \\
& Retrieve (4k) & \mrcolorOne37.65 &   \mrcolorOne\textbf{23.22} & \mrcolorTwo\textbf{21.15} & \mrcolorTwo\textbf{18.27} & \mrcolorThree\textbf{52.45} & \mrcolorThree\textbf{42.78} &   \mrcolorFour29.60 &   \mrcolorFour17.97 & \mrcolorFive48.50      & \mrcolorSix37.87 \\
& Long (16k)     &  \mrcolorOne23.53  & \mrcolorOne 15.75    &  \mrcolorTwo14.28 &  \mrcolorTwo 8.79    &  \mrcolorThree43.60  & \mrcolorThree29.26      &  \mrcolorFour28.65   &     \mrcolorFour \textbf{18.15} &  \mrcolorFive51.00   & \mrcolorSix32.21         \\
& Segment\textsuperscript{+} (4k)      &   \mrcolorOne\textbf{51.62} & \mrcolorOne17.00 &   \mrcolorTwo16.43 &   \mrcolorTwo11.83&   \mrcolorThree42.05 &   \mrcolorThree31.68 &   \mrcolorFour\textbf{34.70}  & \mrcolorFour14.75 &   \mrcolorFive\textbf{52.50}  &   \mrcolorSix\textbf{39.46} \\

\cdashlinelr{1-12}
\multirow{4}{*}{Mistral-7B} & Vanilla (4k)  & \mrcolorOne50.70 &   \mrcolorOne21.84 & \mrcolorTwo17.13 & \mrcolorTwo11.93 & \mrcolorThree43.73 & \mrcolorThree25.15 & \mrcolorFour18.55 & \mrcolorFour12.53 &  \mrcolorFive55.50         & \mrcolorSix37.12 \\
& Retrieve (4k) & \mrcolorOne51.62 &   \mrcolorOne21.84 & \mrcolorTwo23.52 & \mrcolorTwo14.89 & \mrcolorThree56.83 & \mrcolorThree33.04 & \mrcolorFour29.50  & \mrcolorFour16.83 &  \mrcolorFive61.50         & \mrcolorSix44.59 \\
& Long (16k)     &  \mrcolorOne\textbf{59.73} & \mrcolorOne \textbf{27.09}    &  \mrcolorTwo24.55 &  \mrcolorTwo  \textbf{17.03}   &  \mrcolorThree\textbf{63.58}  &  \mrcolorThree \textbf{35.10}    &  \mrcolorFour30.70   & \mrcolorFour \textbf{19.38}& \mrcolorFive\textbf{65.00}    &   \mrcolorSix\textbf{48.71}            \\
& Segment\textsuperscript{+} (4k)      &   \mrcolorOne54.83 & \mrcolorOne17.00 &   \mrcolorTwo\textbf{26.98}  &   \mrcolorTwo12.62 &   \mrcolorThree56.70 &   \mrcolorThree32.19 &   \mrcolorFour\textbf{37.32} &   \mrcolorFour14.75 &    \mrcolorFive59.00       & \mrcolorSix46.97 \\

\bottomrule
\end{tabular}%
\caption{Comparison of main results across various models and datasets. The context window in parentheses refers to the working window size limited for comparison. The highest score in each column is highlighted in \textbf{bold}. Scores are measured using the F1 metric for the `Auto' column, while the `GPT4' column reflects the evaluation scores of GPT-4. Segment\textsuperscript{+} achieves the highest performance relative to other baselines, with the exception of Mistral-7B, which shows comparable performance in settings with the 16k-contexts model. It particularly outperforms agent-like baselines such as MemGPT and Pearl.}

\label{tab:my-table}
\end{table*}
}

Understanding long documents has long been a common research issue in the NLP field, posing challenges due to the increasing length of texts and the complexities involved in comprehensive reasoning.

\paragraph{Benchmarks} 
The datasets utilized in our study are extracted from two notable benchmarks in document understanding, Scrolls~\cite{shaham-etal-2022-scrolls} and Longbench~\cite{bai2023longbench}, which are specifically designed to rigorously evaluate the capabilities of LMs in processing and reasoning through lengthy texts across diverse domains. The selected datasets include Quality (QLTY) and NarrativeQA (NQA), which focus on storytelling; Qasper, which is tailored to scientific articles; and HotpotQA (HQA) and Musique (MSQ), which are aimed at assessing factual knowledge and multi-hop question answering, akin to Wikipedia sources. The diverse range of source texts and task categories, with sample sizes of 200 for each dataset except for NQA, which comprises 100 samples, ensures a comprehensive and exhaustive evaluation of LMs across varied contexts and document lengths. For further elaboration on each dataset, readers are referred to Appendix ~\ref{app:dataset}, while Table ~\ref{tab:selected_dataset_tokens} provides the token count for each dataset.

In our evaluation process, we implement a combination of automatic metrics (Auto) and GPT-4~\cite{openai2023gpt4} evaluation. The GPT-4 evaluation prompt has been slightly adapted in accordance with the methodology proposed by ~\citet{li2023loogle} to change the original score range to 0-100, as detailed in the Appendix~\ref{app:prompt}. We mainly adopt the GPT-4 metrics because the automated metrics focus only on surface-level matching and lack semantic understanding.

\paragraph{Baselines} 
We evaluate a variety of baseline models for processing long documents. Initially, we look at small-context models with a 4k token limit, followed by retrieval methods using the advanced Contriever model~\cite{lei-etal-2023-unsupervised}, also with a 4k token window. For broader contexts, we examined models that handle 16k tokens, which is adequate for most of our experimental data. Our review included ChatGPT (16k)~\cite{openai2022chatgpt}, GPT-4 (128k)~\cite{openai2023gpt4}, Vicuna-7B (4k and 16k versions)~\cite{chiang2023vicuna}, Vicuna-13B (4k and 16k versions)~\cite{chiang2023vicuna}, and Mistral-7B-v0.2 (32k)~\cite{jiang2023mistral}.
We keep the final stage prompt of \method consistent with all the baselines provided by Longbench ~\cite{bai2023longbench}; more details are provided in the appendix~\ref{app:fivedataset}. The temperature for all models is set to 0 for replication purposes.

Motivated by the shared objective of leveraging limited working memory, we contrast our approach with MemGPT~\cite{packer2023memgpt} and implement a script for automatically prompting user responses for answers.
Likewise, for improved reasoning over long documents, we include Pearl~\cite{sun2023pearl} in our comparison. For the action mining process, we use the released resources for QULT and NQA, as they both pertain to the stories domain. We also run this process for other domains where the actions may differ.

Due to the high foundational capabilities of these methods, we tested ChatGPT and GPT-4 according to their respective settings. However, in the case of MemGPT, using ChatGPT almost never yields valid responses; the agent typically awaits user input rather than solving the question, even when task input is provided. Specifically, for the Pearl method, the model is required to generate and execute a plan, which may fail. In such instances, invalid responses are counted as errors.

\begin{figure*}[ht]
\centering
\includegraphics[width=\textwidth]{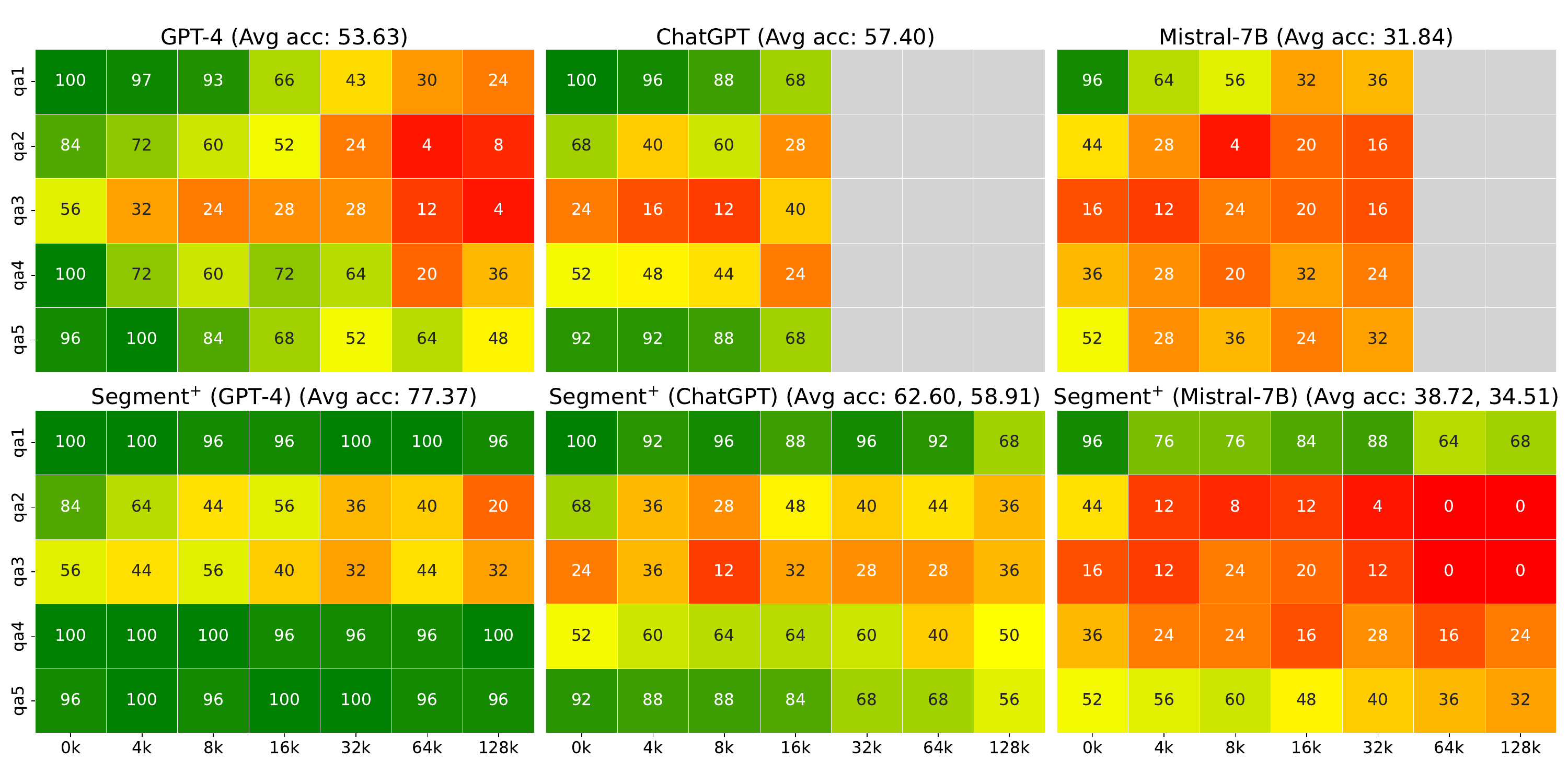} 
\caption{Babilong~\cite{kuratov2024search} Test Performance Comparison. The x-axis represents the length of the input. The y-axis shows the Exact Match (EM) performance on the Babilong task. Results for GPT-4 are taken from Babilong, with each task consisting of 25 items, consistent with the Babilong setting. The average accuracy (Avg acc) for vanilla models and \method (GPT-4) denotes the mean score of all colored cells. However, for \method (ChatGPT) and \method (Mistral-7B), we calculate two average scores: the initial score represents the average over valid contexts for comparison with vanilla models, while the subsequent score indicates the average over all cells. Green indicates higher performance, while red signifies lower performance. \method enhances overall accuracy and maintains stable performance as context length increases.}
\label{fig:example}
\end{figure*}

\paragraph{Main Results} 

Our method, \textbf{\textit{\method}}, exemplifies unparalleled adaptability across LMs of all sizes by segmenting tasks into digestible pieces. This approach empowers smaller models, such as Vicuna-7B and Mistral-7B-v0.2, to excel against a diverse array of benchmarks with remarkable efficiency.

\textbf{The stronger the base model, the greater the performance gain. For larger models, our \method significantly surpasses all comparative baselines.}, delivering a over 20\% performance improvement over vanilla models, which highlights \method's remarkable capability.  
Significantly, \method, alongside Pearl, achieves substantial performance enhancements with GPT-4 compared to ChatGPT. This underscores the fact that agent-based frameworks, when coupled with a meticulously designed reasoning schema and enhanced computational capabilities, can realize notable progress.
This leap forward underscores the critical role of a systematic structure in boosting model performance, especially in the nuanced realm of long document question answering.

\textbf{Nonetheless, the robustness of such a system is equally vital}. MemGPT, when paired with ChatGPT, often fails to respond, terminates abruptly, or excessively depends on human input, undermining reliability. This contrasts sharply with previous agent-based methods that struggled without robust model foundations. Unlike MemGPT, which can stall, awaiting further instructions, \method's well-designed process and schema ensure it remains effective and robust. This demonstrates \method's capability to navigate the complexities of LMs effectively, advocating strongly for its adoption. For smaller models, \method not only achieves a performance increase of over 20\% over vanilla models but also surpasses the performance gains of retrieval models, further demonstrating our method's robustness.

\subsection{Needle-in-a-Haystack Question Answering}
\label{sec:babilong}
Needle-in-a-Haystack~\cite{LLMTest_NeedleInAHaystack2023} has recently become a popular task for testing the processing of long texts. However, we do not choose the original task because it is too artificial. We believe that the reasoning task is more suitable for evaluating long input processing.
We follow \citet{levy2024task}, who reported a decline in the reasoning performance of LMs as the input size increases across various tasks in similar settings.

\paragraph{Benchmark}
We adapt the Babilong benchmark~\cite{kuratov2024search}, which poses a significant challenge as it requires the model to extract and process distributed facts within extensive texts, culminating in reasoning to arrive at a final answer. This tests the model's ability not only to find relevant information but also to reason over it. In line with the main experimental settings of Babilong, we select tasks from qa1 to qa5 for evaluation, using the `Evidence' part of the collected notes for information processing. The context ranges from `0k' to `128k' tokens, where `0k' indicates a context-free environment containing only the given facts, and `4k' to `128k' denotes contexts that include these facts along with noisy data. Given that the output format is fixed, we employ an exact match approach to measure accuracy (\%).

\paragraph{Baselines}
Following the experimental setting of Babilong, we have chosen GPT-4 (128k)~\cite{openai2023gpt4}, ChatGPT (16k)~\cite{openai2022chatgpt}, and Mistral-7B-v0.2 (32k)~\cite{jiang2023mistral} for comparison.
We keep the final stage prompt of \method consistent with all the baselines provided by Babilong. The temperature for all models is set to 0 for replication purposes.

\paragraph{Results} 
\method demonstrates superior performance on the Babilong tasks compared to all models, showcasing its robust anti-noise capabilities and efficient information gathering and reasoning. 
Notably, with \method, ChatGPT even surpassed the performance of vanilla GPT-4, achieving a 5.28\% higher accuracy. 
Furthermore, the stronger the base model, the greater the performance gains observed. This is likely due to the base model's enhanced capabilities, which better leverage the \method strategy to achieve improved performance.

The performance of \method on Babilong tasks remains relatively stable as the length of the input text increases. This stability is due to our method's ability to decompose the task's complexity during long input processing, allowing the model to process only a small piece of text at a time.

\subsection{Ablations}
Do filtered and structured information play crucial roles in the effectiveness of our framework?
To analyze this question, we establish three ablation baselines. First, to examine the effect of information filtering, we eliminate the labeling process and rely solely on the \method, using all chunks to generate the final answer. Second, to assess the impact of structured information, we disregard the structured format and simply aggregate filtered answers from various chunks to formulate the final response in free text. Finally, to evaluate the system without both the structure and filter, we run the chunk and merge algorithm.

\paragraph{Results} Experiment results indicate that both design elements contribute to performance. Additionally, for Vicuna-7b, the filtering module plays a more important role. 
Furthermore, on the Musique dataset, which tests reasoning with challenging distractors and avoids shortcuts, \method demonstrates efficiency on complex multi-hop questions(see Appendix~\ref{app:musique}).
Our methodology substantially improves task performance over vanilla models due to two key factors: \textbf{\textit{1)}} the filtration of pertinent information, and \textbf{\textit{2)}} a structured process for integrating information. This approach not only expands the context window but also processes content more efficiently.

\begin{figure}[t!]
    \centering
    \includegraphics[width=\linewidth]{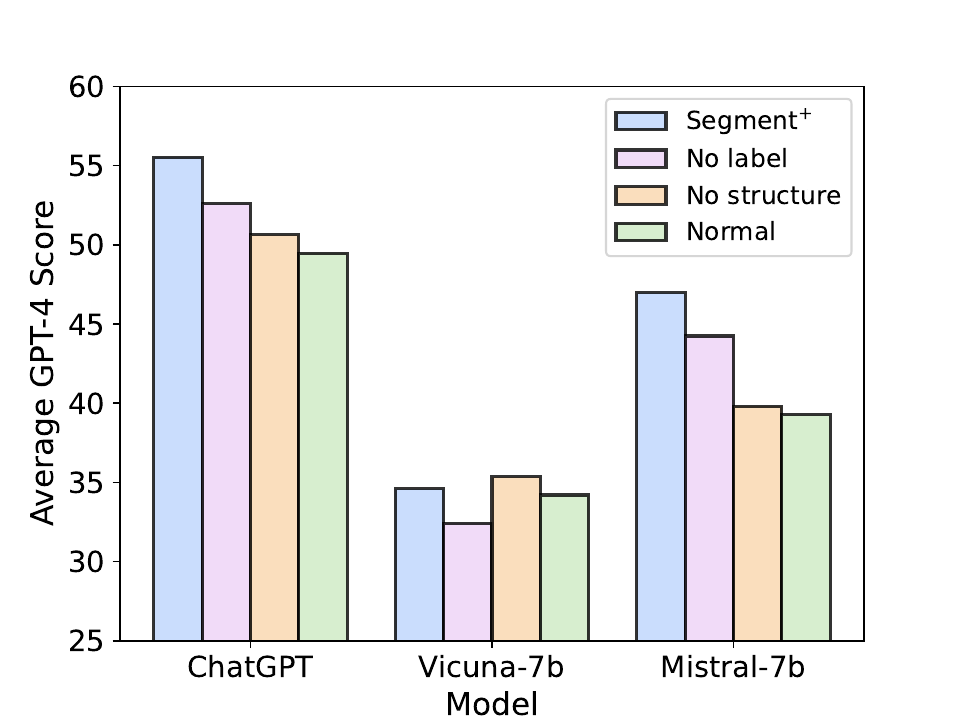}
    \caption{Ablation study results. `No Label' refers to the condition without information filtering, `No Structure' refers to the absence of a structured prompt, and `Normal' indicates the model operates without both filtering and structured prompts. The results demonstrate that both design elements contribute to the final performance.
}
    \label{fig:ablation}
\end{figure}



\subsection{Segment Size Analysis}
Given our use of the segmenting method, it is essential to analyze whether different segment sizes influence the performance of our approach. We examine performance variations as segment size increases from 1000 to 3000 tokens, in increments of 500. Additionally, we average the results across all five datasets in long document processing tasks.

\paragraph{Results}
The average performance of \method across different segment sizes appears stable, with higher performance at the 3000 segment size. A larger segment size brings these advantages: \textbf{\textit{1)}} The information within one segment is more complete, reducing the model's pressure to integrate information. \textbf{\textit{2)}} Fewer segments lead to faster prediction speeds, improving efficiency.

\begin{figure}[t!]
    \centering
    \includegraphics[width=\linewidth]{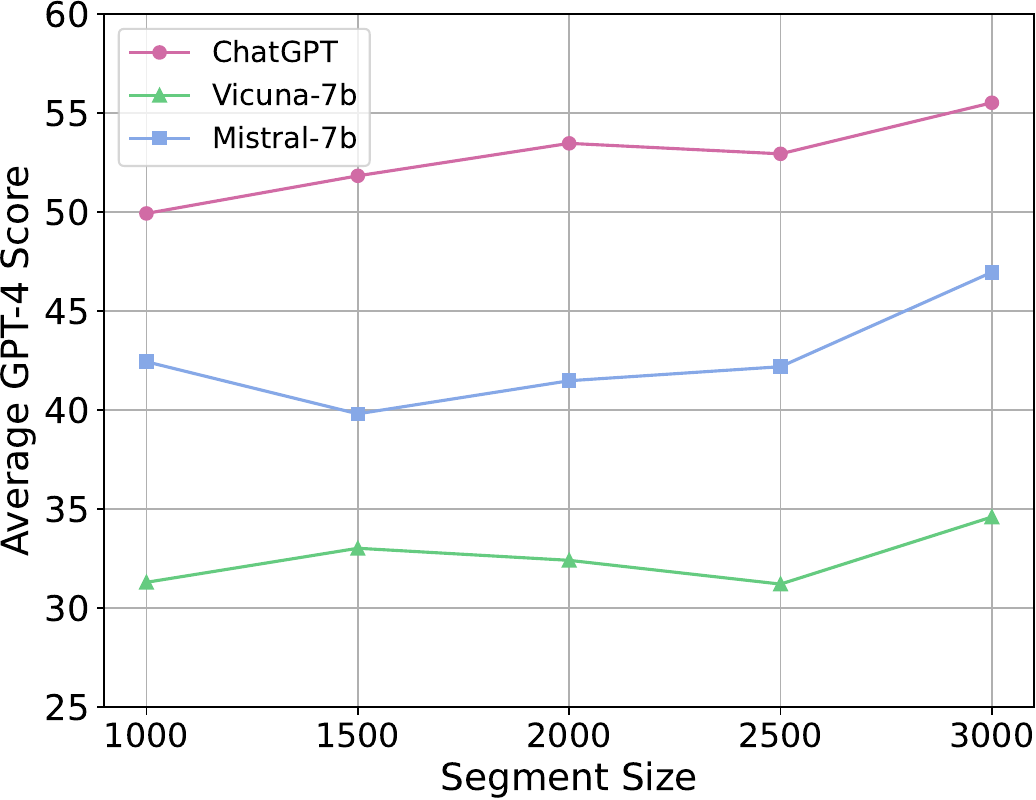}
    \caption{Segment Size Results. The average performance in long document question-answering tasks remains stable across different segment sizes, with optimal results achieved at a segment size of 3000.
}
    \label{fig:chucksize}
\end{figure}

\section{Conclusion}
\label{sec:conclusion}
In this paper, we introduce Segment\textsuperscript{+}, a simple yet effective plug-and-play methodology designed to augment the processing of long inputs within limited context windows, leveraging structured information flow control motivated by data characteristics and filtering mechanisms. Our extensive experiments and analyses substantiate that Segment\textsuperscript{+} significantly enhances performance in long document question answering and noisy text processing, thereby illustrating its broad applicability across diverse domains in this field. When compared to agent-based methods, Segment\textsuperscript{+} not only achieves superior performance but also exhibits greater stability. Furthermore, this information control schema holds potential for broader applications in scenarios requiring long input processing, such as in agent memory management and video information processing.

\section*{Limitations}
\label{sec:limitation}
First, our method is primarily focused on document input processing; it cannot be directly applied to more complex structured texts such as code or text-based games. However, we believe that the underlying concept can be adapted to design specific structures for these scenarios. Second, we notice that \method is more effective when applied to stronger models. This may be due to the strong models' good instruction-following abilities, allowing them to adhere well to our schema design, and their robustness, providing better resistance to textual noise and enhanced reasoning capabilities.

\section*{Ethics Statement}
This paper introduces a novel framework for long-context processing, evaluated on publicly available datasets such as Scrolls, Longbench, and Babilong; therefore, no specific ethical considerations are addressed.

\section*{Acknowledgement}
We are deeply grateful to Yikai Zhang, Jian Xie, and Siyu Yuan from Fudan University for their insightful suggestions and thoughtful discussions, which greatly contributed to this work. This work was supported by Ant Group Research Fund.

\bibliography{custom}
\bibliographystyle{acl_natbib}

\clearpage
\begin{appendix}
\label{sec:appendix}
\section{Dataset Description}
\label{app:dataset}
Datasets from the validation set of Scolls~\cite{shaham-etal-2022-scrolls}:
\begin{itemize}
    \item \textbf{QUALITY~\cite{pang-etal-2022-quality}}: Deep textual engagement is required for this concise yet difficult multiple-choice question set about stories and movies. There is one correct answer for each question, which is meant to assess thorough comprehension rather than just skimming or reading an overview. 
    
    \item \textbf{NarrativeQA~\cite{kocisky-etal-2018-narrativeqa}}: The movie scripts that make up this dataset were gathered from different websites. It requires models to produce free-text responses to predetermined queries. The task pushes the models to go beyond simple pattern matching or salience cues, forcing them to engage in deep reasoning over lengthy scripts or books.
    
    \item \textbf{Qasper~\cite{dasigi-etal-2021-dataset}}: This dataset, which is distinguished by its logical and well-structured content, focuses on providing answers to questions within the context of academic research papers.
    Because the question collectors were not exposed to the entire content of the papers, some questions might go into great detail or even be unanswerable.

\end{itemize}
\par Datasets from the test set of Longbench~\cite{bai2023longbench}:
\begin{itemize}
    \item \textbf{HotpotQA~\cite{yang-etal-2018-hotpotqa}}: The dataset under focus is a question-answering set derived from Wikipedia that requires multi-hop reasoning over various passage segments.  
    \item \textbf{Musique~\cite{trivedi-etal-2022-musique}}: With an emphasis on minimizing train-test leakage, this multi-hop reasoning dataset is designed to get around the shortcuts found in datasets of a similar nature. 
    In order to increase the difficulty and put a model's reasoning skills to the test, it also presents increasingly complicated distractor contexts.

\end{itemize}

\section{GPT-4 Evaluation Prompt}
\label{app:prompt}
\mdfsetup{
  roundcorner=10pt,
  backgroundcolor=gray!15, 
  linewidth=0pt 
}
The evaluation prompt of the Qasper, HQA, MSQ, and NQA datasets was conducted using a prompt structure based on the methodology described in ~\cite{li2023loogle}.
\begin{mdframed}
There is a ground truth answer to a question and an auto-generated answer. Please compare the generated answer with the ground truth and evaluate the generated answer from the perspectives of information completeness, consistency, fluency, and grammar by giving a score within the range of 0 to 100. 

Question = \textit{question}

Groundtruth answer = \textit{answer}

Generated answer = \textit{prediction}

Score = 
\end{mdframed}

The evaluation prompt of the Quality dataset is shown below.

\begin{mdframed}
Give you a 4-choice question and its correct answer (only one choice is correct). You need to check whether the model prediction answer is correct or not. Let's do it step by step.

1. You should carefully read the first and the last sentence of the model prediction. If more than one choice is mentioned in the prediction, you should read the whole prediction carefully and figure out the final predicted answer.

2. Turn the answer into (A), (B), (C) or (D).

3. If the correct answer is choosed and is the only choosed answer, then you can say `true'. If the model give false, none or multi-answers, you should give `false'.

Question: \textit{question}

Correct answer: \textit{answer}

Model prediction: 
\textit{prediction}

Model predicted options:

Correct option:

Evaluation:
\end{mdframed}

\section{Long Document Question Answering Prompt}
\label{app:fivedataset}
The query prompt for HotpotQA, which is slightly modified for other datasets to suit task descriptions.
\begin{mdframed}
You are provided with a segment from a long document along with a question related to this document.

Segment Content:
{segment}

---

Question:
{question}

---

Your task:
Evaluate the provided segment against the question to identify and categorize information into two distinct types: "Evidence" and "Reasoning". Your assessment and categorization should adhere to the following guidelines:

Guidelines for Note-Writing:

Your note should be meticulously structured into two main parts: Evidence and Reasoning, following these guidelines:

      - Evidence:
      
      1. Extract key sentences or descriptions from the segment that are pertinent to the question, with a focus on specific details such as numbers, relevant words, and other significant elements.
      
            (1) Include content that directly relates to the question, providing a straightforward answer.
            
            (2) Also include content that may not directly answer the question but is valuable for answering it when combined with information from other segments. For instance, for questions about someone's birthplace, include all mentioned birthplaces for potential matching in later analysis. Similarly, if the question involves several events but this segment only contains information about one event, you should include it.
            
      2. Accurately quote the directly related sentences to present clear and unambiguous evidence.

      - Reasoning:
      
      1. Analyze the question and any sub-questions, offering answers, summaries, interpretations, or any relevant commentary to deepen the understanding of the question.
          
The note should be formatted in JSON as follows:

\{
  "Evidence": "Your evidence content here",
  
  "Reasoning": "Your reasoning content here"\}
\end{mdframed}

The merge prompt for HotpotQA, which is the same for other datasets to suit task descriptions.
\begin{mdframed}
You are presented with the collection of information from various parts of a lengthy document, along with a specific query that requires a response. The collected information is clearly divided into two parts: Evidence and Reasoning. The Evidence comes from original content of the article, the Reasoning is the model's interpretation based on this evidence.

Collected information:
{notes}

---

Question:
{question}

---

Detailed Instructions:
Process the information from these notes in two separate parts: merge the Evidence sections together and then merge the Reasoning sections.

1. Evidence Synthesis: Examine the Evidence section closely, preserving original content that could possibly help answer the query. Aim to retain as much information as possible without omission.

2. Reasoning Enhancement: Ensure the reasoning is clear, well-structured, and concisely addresses the query within 1-2 sentences.

Upon completing your analysis, update the collected information in the following JSON format:

\{
  "Evidence": "Your evidence content here",
  
  "Reasoning": "Your reasoning content here"
  \}
\end{mdframed}

\section{Ablation Study on Musique Dataset}
\label{app:musique}
\begin{table}[h!]
    \centering
    \resizebox{\linewidth}{!}{
    \begin{tabular}{lcccc}
        \toprule
        \textbf{Model} & \textbf{Segment+} & \textbf{No Label} & \textbf{No Structure} & \textbf{Normal} \\
        \midrule
        ChatGPT      & 67.96    & 35.325   & 33.65    & 28.55  \\
        Vicuna-7b    & 44.42    & 14.35    & 26.20    & 20.70  \\
        Mistral-7b   & 52.35    & 24.02    & 22.52    & 16.38  \\
        \bottomrule
    \end{tabular}
    }
    \caption{Evaluation of ablation baselines on the Musique dataset using GPT-4 scores.}
    \label{tab:musique_comparison}
\end{table}

\end{appendix}

\end{document}